%% file: acl_latex.tex
\pdfoutput=1

\documentclass[11pt]{article}

\usepackage[preprint]{acl}

\usepackage{times}
\usepackage{latexsym}
\usepackage{multirow}
\usepackage[T1]{fontenc}

\usepackage[utf8]{inputenc}
\usepackage{algorithm}
\usepackage{algpseudocode}
\usepackage[inline]{enumitem}
\usepackage{microtype}

\usepackage{inconsolata}

\usepackage{soul}
\usepackage{colortbl,xcolor}
\usepackage{graphicx}
\usepackage{tabularx,arydshln}
\usepackage{booktabs}
\usepackage{amsmath}
\title{Decompose and Compare Consistency: Measuring VLMs' Answer Reliability via Task-Decomposition Consistency Comparison
}

\author{
   Qian Yang$^{1,2}$, Weixiang Yan$^{3}$, Aishwarya Agrawal$^{1,2,4}$
    \\
  $^1$ Mila - Québec AI Institute
  $^2$ Université de Montréal \\
     $^3$ University of California, Santa Barbara 
   $^4$ Canada CIFAR AI Chair 
   \\
    \normalsize {\tt\small qian.yang@mila.quebec}~~~~ {\tt\small weixiangyan@ucsb.edu}~~~~ {\tt\small aishwarya.agrawal@mila.quebec}
}

\begin{document}
\maketitle

\input{Sections/0_abstract}
\input{Sections/1_Introduction}
\input{Sections/2_RelatedWork}
\input{Sections/3_Method}
\input{Sections/4_Experiments}
\input{Sections/5_conclusion}
\clearpage


\section*{Acknowledgement}
We are grateful to Mila’s IDT team for their technical support with the computational infrastructure.
We thank Rabiul Awal for his constructive feedback. 
During this project, Aishwarya Agrawal was supported by the Canada CIFAR AI Chair award.

\section*{Limitations}
Our experiments demonstrate that consistency comparison based on task decomposition can better measure the reliability of VLM answers. However, there are several limitations to our current study:
\textit{Decomposition Performance}: The effectiveness of our framework is influenced by the performance of the decomposition process. Currently, we have not fully explored the optimization and impact of different decomposition strategies for reliability measurement.
\textit{Multi-Agent Consistency Comparison}: We tested decomposition with only one LLM for the multi-agent part. Conducting more experiments with various LLMs will help assess the generalization and robustness of our framework.
Future work will address these limitations to validate and enhance the generalization of our proposed method.

\bibliography{custom}
\clearpage
\appendix
\input{Sections/6_Appendix}

\end{document}

%% file: Sections/0_abstract.tex
\begin{abstract}
Despite tremendous advancements, current state-of-the-art Vision-Language Models (VLMs) are still far from perfect.
They tend to hallucinate and may generate biased responses.
In such circumstances, having a way to assess the reliability of a given response generated by a VLM is quite useful. 
Existing methods, such as estimating uncertainty using answer likelihoods or prompt-based confidence generation, often suffer from overconfidence. Other methods use self-consistency comparison but are affected by confirmation biases.
To alleviate these, we propose \textbf{De}compose and \textbf{C}ompare \textbf{C}onsistency (\texttt{DeCC}) for reliability measurement. 
By comparing the consistency between the direct answer generated using the VLM's internal reasoning process, and the indirect answers obtained by decomposing the question into sub-questions and reasoning over the sub-answers produced by the VLM, \texttt{DeCC} measures the reliability of VLM's direct answer.
Experiments across six vision-language tasks with three VLMs show \texttt{DeCC}'s reliability estimation achieves better correlation with task accuracy compared to the existing methods. 
{The code is publicly available at}
\url{https://github.com/MyLittleChange/DeCC}.
\end{abstract}

%% file: Sections/1_Introduction.tex
\section{Introduction}
 
Automatic measurement of reliability of responses generated by AI systems such as vision-language models (VLMs) is useful for deciding whether to trust a response or not, which in turn is necessary to build secure systems and enable further improvements~\cite{varshney2023post}.
Existing reliability estimation methods often estimate the model's uncertainty using answer likelihoods or prompt the model to generate a confidence value~\cite{xiong2024can, tian-etal-2023-just, mielke-etal-2022-reducing}.
These methods often fail to correlate well with task accuracy because models are not well-calibrated and tend to be overconfident~\cite{chen-etal-2023-close}.
Other methods attempt to incorporate calibrated confidence generation as a training goal~\cite{lin2022teaching, ye-durrett-2022-explanations, oh2024towards}, but retraining the model is inefficient and even impractical for measuring the reliability of multiple VLMs or closed-source models.
Some works use self-consistency to measure reliability by comparing the consistency among multiple generated answers~\cite{wang2022self, chen2024two, chen2023inside}, but self-consistency might suffer from confirmation biases~\cite{feng2024dont}.

\begin{figure}[tb]
  \centering
  \includegraphics[width=\linewidth]{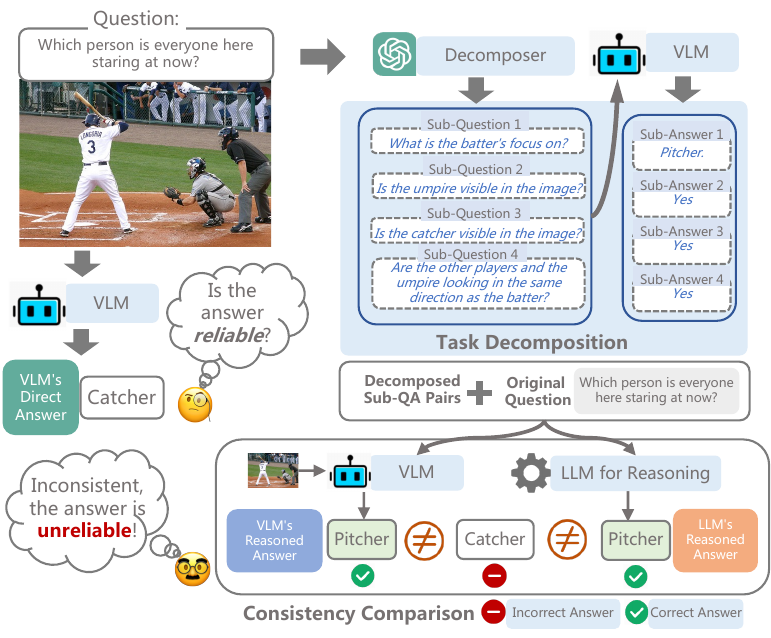}
  \caption{\texttt{DeCC} begins by decomposing the question into multiple sub-questions. The candidate VLM answers these sub-questions, creating sub-QA pairs. Both the candidate VLM and an LLM independently reason over these pairs to derive reasoned answers. We then compare the direct answer with the reasoned answers to assess reliability. We also explore how different consistency comparison settings impact \texttt{DeCC}'s effectiveness.}
  \label{fig:decomp}
   \vspace{-10pt}
\end{figure}

To better measure VLMs' answer reliability, we propose a method called \textbf{De}compose and \textbf{C}ompare \textbf{C}onsistency (\texttt{DeCC}). 
As shown in Fig~\ref{fig:decomp}, we first decompose the original question into several sub-questions. 
The candidate VLM then answers these sub-questions, generating a sequence of sub-QA pairs. 
We use both the candidate VLM and a separate LLM, acting as two independent agents, to reason over the sub-QA pairs and obtain their respective reasoned answers. We then compare the consistency between these reasoned answers and the answer generated directly by the VLM to measure the reliability of the VLM's direct answer.
Using the candidate VLM to reason over sub-QA pairs provides insights into how robustly the VLM understands the question. However, such self-consistency can sometimes introduce confirmation biases~\cite{feng2024dont}. Thus, we also employ an LLM to reason over the sub-QA pairs separately. 
We test both single-agent and multi-agent settings.
For the single-agent setting, we use the consistency between the direct answer and one of the agent's reasoned answers to determine reliability. For the multi-agent setting, we combine the consistency check results from both agents to determine if the answer is reliable, unreliable, or requires further information for measurement.
We assume that if the VLM understands the question well and conducts reliable reasoning, a conflict is less likely to occur between its direct answer, derived from its internal reasoning process, and the decomposed answer, derived from an external reasoning process. 
We evaluate \texttt{DeCC} on six vision-language tasks using three different state-of-the-art VLMs. Experimental results demonstrate that \texttt{DeCC}, which is both model-agnostic and task-agnostic, exhibits a higher correlation with the VLMs' task accuracy compared to the existing methods. Additionally, we observe that the effectiveness of different consistency comparison settings is correlated with the candidate VLM's capabilities.

%% file: Sections/2_RelatedWork.tex
\section{Related Work}

Existing methods use uncertainty-based metrics for reliability measurement, 
such as setting a reliability threshold on answer likelihoods~\cite{pereyra2017regularizing, geifman2017selective, whitehead2022reliable}, or prompting the model to generate a confidence value~\cite{xiong2024can, tian-etal-2023-just, li2024few, mielke-etal-2022-reducing}.
However, uncertainty-based metrics often lead to overconfidence since confidence calibration is not a training goal~\cite{chen-etal-2023-close}.
But retraining models to generate calibrated confidence~\cite{oh2024towards, lin2022teaching, zhang2023r} is impractical for evaluating multiple VLMs.
Self-consistency methods generate multiple responses to assess reliability~\cite{wang2022self, chen2024two, chen2023inside} but suffer from confirmation biases~\cite{huang2024large, xie2024adaptive}. 
Multi-agent collaboration can mitigate this. \citet{feng2024dont} use multiple LLMs to interact in cooperative and competitive settings to evaluate reliability. \citet{srinivasan2024selective} use LLMs to generate related questions about the image and use high-confidence QA pairs as premises, with the original QA as the hypothesis, to determine reliability.
Our approach differs by decomposing the question into simpler sub-questions. 
We also conduct extensive experiments to explore the effectiveness of different consistency comparison settings on reliability measurement.

%% file: Sections/3_Method.tex
\section{Method}

For a question \( Q \), an image \( I \), and an answer \( A \) from a candidate VLM, \texttt{DeCC} obtains {a binary Reliability score $\mathcal{R}$ indicating whether \( A \) is Reliable.} As shown in Fig~\ref{fig:decomp}, \texttt{DeCC} contains two components: Task Decomposition and Consistency Comparison.

\subsection{Task Decomposition}

First, the decomposer, which could be any VLM, decomposes the question \( Q \) into a sequence of sub-questions conditioned on \( I \). 
The candidate VLM then answers these sub-questions, resulting in a sequence of sub-QA pairs. Next, the candidate VLM and a separate LLM, acting as two independent agents, reason over the sub-QA pairs and \( Q \), yielding {VLM's reasoned answer \( A_{V}^{'} \) and LLM's reasoned answer \( A_{L}^{'} \).}
To enhance robustness, we also experiment with a two-iteration decomposition process. In the second iteration, sub-QA pairs from the first iteration, along with \( Q \) and \( I \), are used to guide the decomposer in generating additional sub-questions. 
The candidate VLM answers these new sub-questions, conditioned on \( I \) and previous sub-QA pairs, resulting in new sub-QA pairs containing more information. Finally, both agents reason over all sub-QA pairs from both iterations to {provide their updated reasoned answers, \( A_{V}^{''} \) and \( A_{L}^{''} \)} \footnote{{Single quotation mark (\( {'} \)) annotates the first iteration and double quotation mark (\( {''} \)) annotates the second iteration.}}.

\subsection{Consistency Comparison}
\label{sec:consistency_compare}
We explore both single-agent and multi-agent settings for consistency comparison to obtain \( \mathcal{R} \).

\noindent
\textbf{Single-Agent} We compare the VLM's direct answer \( A \) with either the VLM's reasoned answer \( A_{V}^{'} \) (\textit{VLM Agent Consistency}) or the LLM's reasoned answer \( A_{L}^{'} \) (\textit{LLM Agent Consistency}) and obtain:
\[ \mathcal{R} = 
\begin{cases} 
1, & \text{if } A^{'} \text{ is consistent with } A \\
0, & \text{otherwise}
\end{cases}
\]

We check if \( A^{'} \) = \( A \) to determine the consistency.
For two-iteration decomposition, we compare \( A \) with \( A_{V}^{''} \) and \( A_{L}^{''} \) to obtain \( \mathcal{R} \) in a similar way.

\begin{figure}[tb]
  \centering
  \includegraphics[width=0.95\linewidth]{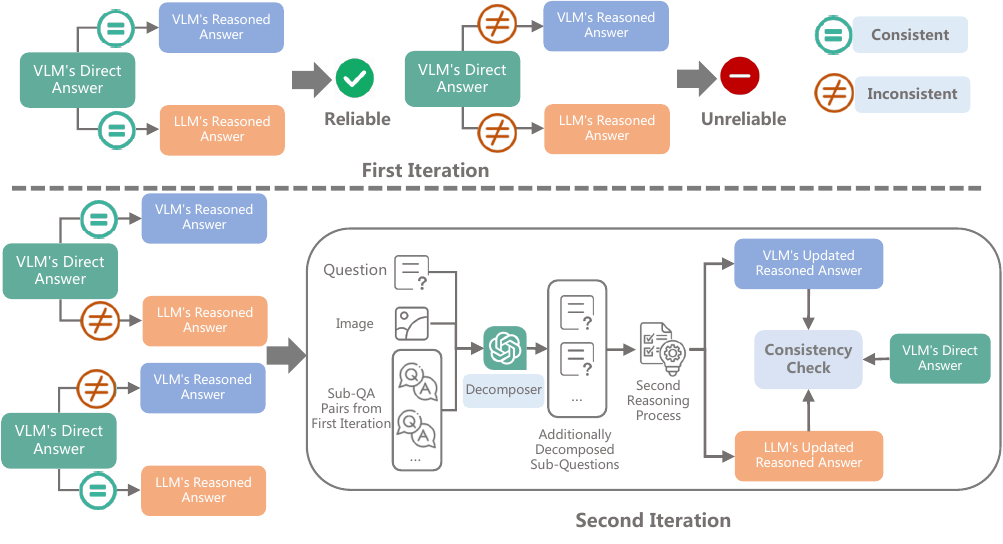}
  \caption{Illustration of Multi-Agent Consistency Comparison. \textit{Top:} When both agents' reasoned answers are either consistent or inconsistent with the VLM's direct answer, we directly determine the Reliability. \textit{Bottom:} If there is a contradiction in consistency check results, we proceed to the second-iteration consistency checks.   }
  \label{fig:consistency}
  \vspace{-10pt}
\end{figure}

\noindent
\textbf{Multi-Agent} As shown in Fig~\ref{fig:consistency}, we first conduct consistency checks of  \( A \) with \( A_{V}^{'} \) and \( A_{L}^{'} \) and obtain \( Cons_{V}^{'} \) (consistency between \( A \) and \( A_{V}^{'} \)) and \( Cons_{L}^{'} \) (consistency between \( A \) and \( A_{L}^{'} \)). If \( Cons_{V}^{'} = Cons_{L}^{'} \), we assign \( \mathcal{R} = Cons_{V}^{'} \). 
If \( Cons_{V}^{'} \neq Cons_{L}^{'} \), we proceed to the second-iteration consistency checks, where we compare updated reasoned answers \( A_{V}^{''} \) and \( A_{L}^{''} \) with \( A \), obtaining \( Cons_{V}^{''} \) and \( Cons_{L}^{''} \). We assign \( \mathcal{R} \) as:
\[
\mathcal{R} = 
\begin{cases} 
Cons_{V}^{''}, & \text{if } Cons_{V}^{''} = Cons_{L}^{''} \\
Cons_{L}^{''}, & {\text{if } Cons_{V}^{'} = Cons_{V}^{''} \text{ and }} \\
& \quad Cons_{L}^{'} = Cons_{L}^{''} \\
Cons_{V}^{''}, & \text{if } Cons_{V}^{'} \neq Cons_{V}^{''} \text{ and } \\
& \quad Cons_{L}^{'} \neq Cons_{L}^{''}
\end{cases}
\]
(1) The first scenario indicates that the consistency check outcome for one of the agents has changed from the first iteration, leading to the same consistency check outcomes between the two agents.
(2) The second scenario indicates that both agents show strong confidence in their respective consistencies with respect to the direct answer. We trust the LLM's consistency check, as it provides a more objective assessment, Relying solely on textual decomposition information, whereas the VLM might suffer from its inherent biases towards certain responses.
(3) The third scenario indicates that the second-iteration decomposition provides additional information, influencing both agents' reasoning and changing their consistency with respect to the direct answer. We trust the VLM’s consistency check outcome, as 
VLM is less likely to change its response due to its inherent biases, whereas the LLM's response is more likely to change since it is operating under incomplete information (lack of image). So a change in VLM's response indicates it potentially overcame its biases with additional sub-QA pairs.
See Appendix for Algorithm
~\ref{alg:multi_agent_consistency}.

%% file: Sections/4_Experiments.tex
\section{Experiments}
\input{Tables/LLaVA_IDE_Intern_Results}

\subsection{Datasets}
{We conduct experiments on six VQA tasks: SNLI-VE~\citep{SNLIVE}, VCR~\citep{VCR}, A-OKVQA~\citep{aokvqa}, Winoground~\citep{winoground}, MMMU~\cite{yue2023mmmu}, and MathVista~\cite{lu2024mathvista}. 
More details can be found on Appendix~\ref{sec:datasets}.}

\subsection{Evaluation Metric}
We use the Brier Score (BS)\cite{brier1950verification} to measure the correlation between reliability and task accuracy: \(\text{BS} = \frac{1}{N} \sum_{i=1}^{N} (\mathcal{R}_i - Acc_i)^2\), where \(N\) is the evaluation dataset size, \(\mathcal{R}_i\) is the reliability score for the \(i\)-th answer, and \(Acc_i\) is the accuracy for the \(i\)-th answer. BS ranges between 0 and 1, with lower values indicating better correlation between \(\mathcal{R}\) and \(Acc\).
We also apply \texttt{DeCC} for the selective prediction task where the model abstains from answering when it's response is estimated to be unreliable. To measure \texttt{DeCC} effectiveness at selective prediction we use the Effective Reliability (ER) metric proposed in ~\cite{whitehead2022reliable}. ER captures the trade-off between risk (task accuracy across all answered questions) and coverage (number of questions answered). Both low risk but low coverage and high coverage but high risk lead to low ER.
ER for the \(i\)-th answer is computed as:
\[
ER(A_i) = 
\begin{cases} 
1 & \text{if } \mathcal{R}_i = 1 \text{ and } Acc_i = 1 \\ 
-1 & \text{if } \mathcal{R}_i = 1 \text{ and } Acc_i = 0 \\ 
0 & \text{if } \mathcal{R}_i = 0 \text{ (answer abstention)}
\end{cases}
\] 

\subsection{Existing Methods Used for Comparison}
\textbf{Perplexity of Direct Answer}: Calculate the mean perplexity over tokens of the direct answer and use a threshold to determine reliability. If perplexity exceeds the threshold, \(\mathcal{R}\) is 0 otherwise 1.
\textbf{Generated Numerical Confidence}: Prompt the VLM to generate a confidence value along with the answer, formatted as `Answer: X. Confidence: X\%'. A threshold determines reliability.
\textbf{Generated Linguistic Confidence}: Prompt the VLM to state `I am confident/not confident in this answer.'
\textbf{Self-Consistency based on Paraphrase}: Prompt a VLM to paraphrase the original question into four variations. If $n$ or more paraphrased answers differ from the direct answer, \(\mathcal{R}\) is 0 otherwise 1.
\footnote{We select the best threshold and $n$ for each VLM based on the Brier Score (results in Tables~\ref{Tab:ide_brier_ppl} and~\ref{tab:ide_brier_paraphrase}).}
\input{Tables/IDE_Decoding}

\input{Tables/Question_Type_Analysis}

\subsection{Main Results}
We conduct experiments on six vision-language tasks\footnote{All datasets are multiple-choice except for MMMU and MathVista, whose answers are very short. We use string matching for consistency comparison.}, covering commonsense reasoning, fine-grained compositional reasoning, and science understanding (see Appendix~\ref{sec:datasets} for dataset descriptions).
We evaluate three state-of-the-art VLMs: LLaVA1.5-7B~\cite{liu2023llava}, Idefics2-8B~\cite{idefics2}, and InternVL1.5-25.5B~\cite{internvl} (see Appendix~\ref{sec:implement} for implementation details).
The overall results are shown in Table~\ref{tab:llava_ide_intern_res}.
\texttt{DeCC} achieves the best and second-best mean performance (mean across datasets) on Brier Score and Effective Reliability. \texttt{DeCC} reduces the relative mean Brier Score by 8.7\% on LLaVA, 14.3\% on Idefics2, and 2.9\% on InternVL compared to the best existing methods. 
\texttt{DeCC} also increases relative mean Effective Reliability by 16.5\% on LLaVA, 25.6\% on Idefics2, and 1.7\% on InternVL.
We observe that with increasing VLM size, the performance of most methods improves, suggesting that reliability measurement is correlated with VLMs' capabilities.
For the effectiveness of \texttt{DeCC}'s different consistency comparison settings, we observe an interesting trend:
\begin{enumerate*}[label=(\arabic*)]
\item For weaker VLMs, i.e., LLaVA, LLM Agent Consistency achieves the best performance, likely because VLMs struggle to reason over the sub-QA pairs and suffer from confirmation biases.
\item For stronger VLMs, i.e. Idefics2, Multi-Agent Consistency performs the best suggesting that the VLM and LLM reasoners complement each other. 
\item For the strongest VLMs, i.e. InternVL, VLM Agent Consistency (self-consistency) achieves the best performance, as the VLM can effectively leverage the information contained in sub-QA pairs.
\end{enumerate*}
Overall, the effectiveness of different consistency comparison settings correlates with the candidate VLM's capabilities.

\subsection{Further Analysis}
\paragraph{Decoding Strategy}
The decoding strategy of decomposer can influence the generated sub-questions, which in turn impacts \texttt{DeCC}’s overall performance. 
We conducted experiments using sampling-based decoding with two different temperature settings, $0.8$ and $0.9$, while keeping the nucleus sampling~\cite{Holtzman2020The} probability fixed at $0.9$ for both.
The results 
are presented in Table~\ref{tab:decoding}, using Idefics2-8B as the candidate VLM. Across all decoding strategies and temperature settings, \texttt{DeCC} consistently outperforms the baselines, demonstrating its effectiveness.

\paragraph{Question Type Analysis}
Previous studies have shown that VLMs can exhibit biases toward certain question types. For instance, they respond ``yes'' to over 80\% of queries about non-existent objects, such as in prompts like ``Is there an object in this image?''~\cite{wang2023evaluation}. 
If the decomposed sub-questions retain the same type as the original question, they may inherit the same bias, affecting evaluation reliability.
To investigate this, we conduct the question type analysis to assess whether \texttt{DeCC} is prone to this issue
\footnote{We categorize questions into types using string matching, with categories including: yes/no, color, number, how, why, what/which, when, where, who, and others.}.
Table~\ref{tab:question_type} shows the number of questions and question types per sample across benchmarks. 
\texttt{DeCC} generates at least 2.1 distinct question types per sample for all benchmarks, reducing the impact of VLM biases.

\paragraph{Additional Analysis}

To better understand the workflow of \texttt{DeCC}, we provide some qualitative examples in Appendix~\ref{sec:case_study}.
We also provide the computational costs analysis in Appendix~\ref{sec:comp_analysis} 
to address concerns about the additional computational requirements of \texttt{DeCC} and demonstrate its practical applicability.


%% file: Tables/LLaVA_IDE_Intern_Results.tex
\begin{table*}[ht]
\centering
\resizebox{\textwidth}{!}{\begin{tabular}{l|cc|cc|cc|cc|cc|cc|cc}
\toprule
 \multirow[m]{2}{*}{\textbf{Method}}  & \multicolumn{2}{c|}{\textbf{SNLI}}  &\multicolumn{2}{c|}{\textbf{VCR}} & \multicolumn{2}{c|}{\textbf{A-OKVQA}} & \multicolumn{2}{c|}{\textbf{Wino.}} & \multicolumn{2}{c|}{\textbf{MMMU}} & \multicolumn{2}{c|}{\textbf{MathVista}} & \multicolumn{2}{c}{\textbf{Mean}} \\
 \cmidrule(lr){2-3} \cmidrule(lr){4-5} \cmidrule(lr){6-7} \cmidrule(lr){8-9} \cmidrule(lr){10-11} \cmidrule(lr){12-13} \cmidrule(lr){14-15}
  & {\textbf{BS}{$\downarrow$}}  & {\textbf{ER}{$\uparrow$}} & {\textbf{BS}{$\downarrow$}}  & {\textbf{ER}{$\uparrow$}} & {\textbf{BS}{$\downarrow$}}  & {\textbf{ER}{$\uparrow$}} & {\textbf{BS}{$\downarrow$}}  & {\textbf{ER}{$\uparrow$}} & {\textbf{BS}{$\downarrow$}}  & {\textbf{ER}{$\uparrow$}} & {\textbf{BS}{$\downarrow$}}  & {\textbf{ER}{$\uparrow$}}  & {\textbf{BS}{$\downarrow$}}  & {\textbf{ER}{$\uparrow$}}\\
\midrule
\textbf{LLaVA1.5-7B as Candidate VLM}  &  \multicolumn{2}{c|}{\textit{Acc: 55.0}}  & \multicolumn{2}{c|}{$\textit{Acc: 59.2}$} & \multicolumn{2}{c|}{$\textit{Acc: 67.3}$} & \multicolumn{2}{c|}{$\textit{Acc: 59.6}$} & \multicolumn{2}{c|}{$\textit{Acc: 34.3}$} &\multicolumn{2}{c|}{ $\textit{Acc: 24.5}$} &\multicolumn{2}{c}{$\textit{Acc: 49.9}$}\\
\text{Perplexity of Direct Answer} & 55.7 & 0.7 & 38.2 & 20.4 & 22.8 & 55.0 & 39.3 & 24.3 & 42.4 & -8.2 & \cellcolor{blue!20}\textbf{25.9} & \cellcolor{blue!20}\textbf{-1.4} & 37.4 & 15.1 \\

\text{Generated Numerical Confidence} &66.5 & -32.5 & 40.8 & 18.3 & 22.1 & 55.6 & \cellcolor{blue!10}\underline{28.0} & \cellcolor{blue!10}\underline{44.0} & 67.3 & -35.3 & 75.8 & -51.6 & 50.1 & -0.2 \\

\text{Generated Linguistic Confidence}  &67.5 & -35.0 & 40.2 & 19.6 & 22.6 & 54.8 & \cellcolor{blue!20}\textbf{27.6} & \cellcolor{blue!20}\textbf{44.8} & 69.6 & -39.1 & 77.2 & -54.4 & 50.8 & -1.5 \\

\text{Self-Consistency based on Paraphrase} &38.5 & 17.5 & 32.8 & \cellcolor{blue!10}\underline{25.7} & 19.0 & 59.2 & 40.5 & 23.9 & 39.1 & -5.6 & 35.6 & -11.5 & 34.3 & 18.2\\

\hdashline
\textbf{\texttt{DeCC} } & & & & & & & & & & & & & &\\
\text{VLM Agent Consistency}& 31.9 & 24.5 & 36.4 & 22.2 & \cellcolor{blue!20}\textbf{18.2} & \cellcolor{blue!20}\textbf{59.6} & 35.3 & 28.3 & 52.3 & -18.1 & 46.3 & -21.8 & 36.7 & 15.8\\

\text{VLM Agent Consistency (2 iterations)} & 32.5 & 23.9 & 34.5 & 24.1 & \cellcolor{blue!10}\underline{18.3} & \cellcolor{blue!10}\underline{59.5} & 36.1 & 27.4 & 49.1 & -14.9 & 45.6 & -21.1 & 36.0 & 16.5\\

\text{LLM Agent Consistency}  & 32.0 & 24.4 & 35.9 & 22.7 & 24.5 & 53.3 & 37.5 & 26.0 & \cellcolor{blue!20}\textbf{34.1} & \cellcolor{blue!20}\textbf{0.1} & \cellcolor{blue!10}\underline{30.7} & \cellcolor{blue!10}\underline{-6.2} & 32.4 & 20.1\\

\text{LLM Agent Consistency (2 iterations)} & \cellcolor{blue!20}\textbf{30.6} & \cellcolor{blue!20}\textbf{25.8} & \cellcolor{blue!20}\textbf{32.6} & \cellcolor{blue!20}\textbf{26.0} & 22.3 & 55.5 & 34.6 & 28.9 & 36.8 & -2.6 & 31.0 & -6.5 & \cellcolor{blue!20}\textbf{31.3} & \cellcolor{blue!20}\textbf{21.2}\\

\text{Multi-Agent Consistency (2 iterations)} & \cellcolor{blue!10}\underline{31.5} &  \cellcolor{blue!10}\underline{24.9} &  \cellcolor{blue!10}\underline{33.5} & 25.1 & 20.1 & 57.7 & 34.6 & 28.9 &  \cellcolor{blue!10}\underline{36.4} &  \cellcolor{blue!10}\underline{-2.2} & 32.2 & -7.7 &  \cellcolor{blue!10}\underline{31.4} &  \cellcolor{blue!10}\underline{21.1}\\

\hline
\hline
\textbf{Idefics2-8B as Candidate VLM} & \multicolumn{2}{c|}{$\textit{Acc: 39.3}$}& \multicolumn{2}{c|}{$\textit{Acc: 78.6}$} & \multicolumn{2}{c|}{$\textit{Acc: 83.1}$} & \multicolumn{2}{c|}{$\textit{Acc: 70.0}$} & \multicolumn{2}{c|}{$\textit{Acc: 39.9}$} & \multicolumn{2}{c|}{$\textit{Acc: 48.0}$} & \multicolumn{2}{c}{$\textit{Acc: 59.8}$} \\
\text{Perplexity of Direct Answer} & 59.7 & -20.0 & 34.1 & 28.2 & 19.9 & 63.2 & 29.8 & 43.5 & 40.6 & -1.0 & \cellcolor{blue!10}\underline{30.0} & 15.1 & 35.6 & 21.5 \\
\text{Generated Numerical Confidence} & 40.8 & -0.5 & 37.7 & 25.3 & 36.3 & 46.7 & 25.3 & 49.1 & 67.7 & -43.6 & 49.3 & -1.6 & 42.8 & 12.6 \\
\text{Generated Linguistic Confidence} & 35.0 & -3.1 & 40.2 & 22.1 & 25.2 & 56.6 & 26.8 & 45.6 & 60.4 & -36.3 & 42.4 & 3.5 & 38.3 & 14.7 \\

\text{Self-Consistency based on Paraphrase} & 59.1 & -19.3 & 31.6 & 30.4 & 16.3 & 66.5 & 28.9 & 43.8 & 41.6 & -2.0 & 40.8 & 4.8 & 36.4 & 20.7 \\
\hdashline
\textbf{\texttt{DeCC} } & & & & & & & & & & & & & &\\
\text{VLM Agent Consistency} & 44.9 & -5.2 & \cellcolor{blue!10}\underline{30.5} & \cellcolor{blue!10}\underline{31.6} & \cellcolor{blue!10}\underline{13.9} & \cellcolor{blue!10}\underline{69.2} & \cellcolor{blue!10}\underline{22.6} & \cellcolor{blue!10}\underline{50.4} & 43.9 & -4.4 & \cellcolor{blue!20}\textbf{28.7} & \cellcolor{blue!10}\underline{15.5} & \cellcolor{blue!10}\underline{30.8} & 26.2 \\
\text{VLM Agent Consistency (2 iterations)} & 47.8 & -8.1 & \cellcolor{blue!20}\textbf{29.5} & \cellcolor{blue!20}\textbf{33.1} & \cellcolor{blue!20}\textbf{13.8} & \cellcolor{blue!20}\textbf{69.3} & \cellcolor{blue!20}\textbf{22.3} & \cellcolor{blue!20}\textbf{50.9} & 43.0 & -3.6 & 29.4 & \cellcolor{blue!20}\textbf{15.9} & 31.0 & \cellcolor{blue!10}\underline{26.3} \\
\text{LLM Agent Consistency} & \cellcolor{blue!20}\textbf{34.3} & 5.5 & 37.9 & 24.4 & 26.3 & 56.5 & 35.3 & 38.0 & \cellcolor{blue!20}\textbf{34.2} & \cellcolor{blue!20}\textbf{5.3} & 40.8 & 4.4 & 34.8 & 22.3 \\
\text{LLM Agent Consistency (2 iterations)} & 34.9 & \cellcolor{blue!20}\textbf{6.3} & 34.0 & 25.0 & 24.0 & 61.4 & 32.0 & 39.3 & 35.9 & \cellcolor{blue!10}\underline{5.1} & 34.0 & 11.4 & 32.5 & 24.8 \\
\text{Multi-Agent Consistency} & \cellcolor{blue!10}\underline{34.7} & \cellcolor{blue!10}\underline{5.8} & 33.0 & 27.9 & 19.6 & 65.5 & 29.5 & 44.1 & \cellcolor{blue!10}\underline{35.1} & 5.0 & 31.1 & 13.5 & \cellcolor{blue!20}\textbf{30.5} & \cellcolor{blue!20}\textbf{27.0} \\
\hline
\hline
\textbf{InternVL1.5-25.5B as Candidate VLM}  & \multicolumn{2}{c|}{\textit{Acc: 70.2}} & \multicolumn{2}{c|}{\textit{Acc: 70.5}} & \multicolumn{2}{c|}{\textit{Acc: 88.5}} & \multicolumn{2}{c|}{\textit{Acc: 78.6}} & \multicolumn{2}{c|}{\textit{Acc: 43.7}} & \multicolumn{2}{c|}{\textit{Acc: 56.0}} & \multicolumn{2}{c}{\textit{{Acc: 67.9}}} \\

\text{Perplexity of Direct Answer} & \cellcolor{blue!20}\textbf{28.0} & \cellcolor{blue!20}\textbf{42.2} & \cellcolor{blue!20}\textbf{27.5} & \cellcolor{blue!20}\textbf{43.6} & 12.1 & 76.4 & 24.0 & 56.1 & \cellcolor{blue!10}\underline{37.3} & \cellcolor{blue!10}\underline{6.3} & 36.5 & 18.7 & 27.6 & 40.6 \\
\text{Generated Numerical Confidence} & 37.8 & 30.2 & 42.2 & 21.2 & 21.2 & 62.0 & 19.0 & \cellcolor{blue!20}\textbf{62.1} & 64.6 & -29.4 & 39.6 & 17.6 & 37.4 & 27.3 \\
\text{Generated Linguistic Confidence} & 58.4 & -26.0 & 31.4 & 37.9 & 15.7 & 68.6 & 43.4 & 13.3 & 71.6 & -43.3 & 43.1 & 10.4 & 43.9 & 10.2 \\

\text{Self-Consistency based on Paraphrase} & \cellcolor{blue!10}\underline{30.1} & \cellcolor{blue!10}\underline{40.1} & \cellcolor{blue!10}\underline{28.1} & \cellcolor{blue!10}\underline{43.0} & \cellcolor{blue!20}\textbf{11.0} & \cellcolor{blue!20}\textbf{77.5} & 21.1 & 59.0 & 48.8 & -5.0 & 52.9 & 3.6 & 32.0 & 36.4 \\

\hdashline
\textbf{\texttt{DeCC} } & & & & & & & & & & & & & &\\

\text{VLM Agent Consistency} & 33.2 & 37.0 & 28.3 & 42.8 & 11.9 & 76.6 & \cellcolor{blue!10}\underline{18.9} & 61.3 & 44.9 & -1.2 & \cellcolor{blue!20}\textbf{23.8} & \cellcolor{blue!20}\textbf{31.4} & \cellcolor{blue!20}\textbf{26.8} & \cellcolor{blue!20}\textbf{41.3} \\
\text{VLM Agent Consistency (2 iterations)} & 33.9 & 36.3 & 29.1 & 42.0 & \cellcolor{blue!10}\underline{11.3} & \cellcolor{blue!10}\underline{77.2} & \cellcolor{blue!20}\textbf{18.6} & \cellcolor{blue!10}\underline{61.5} & 44.8 & -1.1 & \cellcolor{blue!10}\underline{24.3} & \cellcolor{blue!10}\underline{30.9} & \cellcolor{blue!10}\underline{27.0} & \cellcolor{blue!10}\underline{41.1} \\
\text{LLM Agent Consistency} & 36.3 & 33.9 & 37.6 & 33.5 & 22.2 & 66.3 & 29.4 & 50.8 & 40.3 & 3.3 & 37.1 & 18.1 & 33.8 & 34.3 \\
\text{LLM Agent Consistency (2 iterations)} & 34.5 & 35.7 & 34.9 & 36.2 & 18.8 & 69.7 & 27.0 & 53.1 & \cellcolor{blue!20}\textbf{36.9} & \cellcolor{blue!20}\textbf{6.8} & 33.3 & 21.9 & 30.9 & 37.2 \\
\text{Multi-Agent Consistency (2 iterations)} & 34.3 & 35.9 & 32.6 & 38.5 & 15.4 & 73.1 & 23.8 & 56.4 & 37.4 & 6.2 & 31.1 & 24.1 & 29.1 & 39.0 \\
\bottomrule
\end{tabular}}
\caption{Measuring Brier Score (\cellcolor{blue!20}\textbf{BS}) and Effective Reliability (\cellcolor{blue!20}\textbf{ER}) for various reliability measurement methods. Best results are in \cellcolor{blue!20}\textbf{bold}. Second-best results are \cellcolor{blue!10}\underline{underlined}. \textit{Acc} represents the task accuracy of the candidate VLM. All scores are in percentage.
\texttt{DeCC} surpasses all baselines in average Brier Score and Effective Reliability.
}
\label{tab:llava_ide_intern_res}

\end{table*}

%% file: Tables/IDE_Decoding.tex
\begin{table*}[h]
\centering
\vspace{-5pt}
\resizebox{0.77\textwidth}{!}{
\begin{tabular}{l|cc|cc|cc|cc}
\toprule
\multirow{2}{*}{\textbf{Method}} & \multicolumn{2}{c|}{\textbf{VCR}} & \multicolumn{2}{c|}{\textbf{AOKVQA}} & \multicolumn{2}{c|}{\textbf{Winoground}} & \multicolumn{2}{c}{\textbf{Mean}} \\
\cmidrule(lr){2-3} \cmidrule(lr){4-5} \cmidrule(lr){6-7} \cmidrule(lr){8-9}

 & {\textbf{BS{$\downarrow$}}} & {\textbf{ER{$\uparrow$}}} & {\textbf{BS{$\downarrow$}}} & {\textbf{ER{$\uparrow$}}} & {\textbf{BS{$\downarrow$}}} & {\textbf{ER{$\uparrow$}}} & {\textbf{BS{$\downarrow$}}} & {\textbf{ER{$\uparrow$}}} \\
\midrule
{Perplexity of Direct Answer} & 34.1 & 28.2 & 19.9 & 63.2 & 29.8 & 43.5 & 27.9 & 45.0 \\ 
{Generated Numerical Confidence} & 37.7 & 25.3 & 36.3 & 46.7 & 25.3 & 49.1 & 33.1 & 40.4 \\ 
{Generated Linguistic Confidence} & 40.2 & 22.1 & 25.2 & 56.6 & 26.8 & 45.6 & 30.7 & 41.4 \\ 
{Self-Consistency based on Paraphrase} & 31.6 & 30.4 & 16.3 & 66.5 & 28.9 & 43.8 & 25.6 & 46.9 \\ 
\hdashline
\textbf{\texttt{DeCC} - Greedy Decoding} & & & & & & & & \\
{VLM Agent Consistency} & \cellcolor{blue!20}\textbf{30.5} & \cellcolor{blue!20}\textbf{31.6} & \cellcolor{blue!10}\underline{13.9} & \cellcolor{blue!10}\underline{69.2} & \cellcolor{blue!20}\textbf{22.6} & \cellcolor{blue!20}\textbf{50.4 }& \cellcolor{blue!20}\textbf{22.3} & \cellcolor{blue!10}\underline{50.4} \\ 
{LLM Agent Consistency} & 37.9 & 24.4 & 26.3 & 56.5 & 35.3 & 38.0 & 33.2 & 39.6 \\ 
{MultiAgent Consistency} & 33.0 & 27.9 & 19.6 & 65.5 & 29.5 & 44.1 & 27.4 & 45.8 \\ 
\hdashline
\textbf{\texttt{DeCC} - Sampling Decoding (Temp = 0.8)} & & & & & & & & \\
{VLM Agent Consistency} & \cellcolor{blue!10}\underline{31.2}  & \cellcolor{blue!10}\underline{31.1}  & \cellcolor{blue!20}\textbf{13.2} & \cellcolor{blue!20}\textbf{69.9} & \cellcolor{blue!10}\underline{22.9} & \cellcolor{blue!10}\underline{50.3} & \cellcolor{blue!10}\underline{22.4} & \cellcolor{blue!20}\textbf{50.5} \\ 
{LLM Agent Consistency} & 38.1  & 24.2 & 24.6 & 58.5  & 35.1 & 38.2 & 32.6 & 40.3 \\ 
{MultiAgent Consistency} & 32.7 & 29.6 & 18.5 & 64.6  & 27.8  & 47.2  & 26.3 & 47.1 \\ 
\hdashline
\textbf{\texttt{DeCC} - Sampling Decoding (Temp = 0.9)} & & & & & & & & \\
{VLM Agent Consistency} & 31.7 & 30.6 & 14.1 & 69.0 & 23.3 & 49.9 & 23.0 & 49.9 \\ 
{LLM Agent Consistency} & 38.6 & 23.7 & 26.8 & 56.3 & 33.8 & 39.5 & 33.0 & 39.9 \\ 
{MultiAgent Consistency} & 33.7 & 28.6 & 20.9 & 63.3 & 28.4 & 44.9 & 27.7 & 45.6 \\ 
\bottomrule
\end{tabular}
}
\caption{Results of difference decoding strategy using Idefics2-8B as the candidate VLM. Best results are in \textbf{bold}. Second-best results are \underline{underlined}. In all decoding methods and temperatures, \texttt{DeCC} outperforms the baselines, demonstrating its effectiveness.}
\label{tab:decoding}
\vspace{-10pt}
\end{table*}

%% file: Tables/Question_Type_Analysis.tex
\begin{table}[ht]
    \centering
    \resizebox{0.8\linewidth}{!}{
    \begin{tabular}{lcc}
        \toprule
        \multirow{2}{*}{\textbf{Benchmark}} & \textbf{Questions} & \textbf{Question Types} \\
         & \textbf{per Sample} & \textbf{per Sample} \\
        \hline
        SNLI & 4.1 & 2.1 \\
        VCR & 5.2 & 2.2 \\
        AOKVQA & 4.4 & 2.1 \\
        Winoground & 4.3 & 2.5 \\
        MMMU & 5.9 & 2.4 \\
        MathVista & 4.4 & 2.3 \\
        \bottomrule
    \end{tabular}}
    \caption{Overview of the number of questions and question types per sample across different benchmarks.}
    \label{tab:question_type}
    \vspace{-10pt}
\end{table}

%% file: Sections/5_conclusion.tex
\section{Conclusion}
We use consistency comparison based on task decomposition for measuring VLMs answer reliability. 
By decomposing complex questions into simpler sub-questions, we achieve more accurate and robust reliability estimation. 
We find the performance of reliability measurement and the effectiveness of different consistency comparison settings correlate with candidate VLM's capabilities.

%% file: Sections/6_Appendix.tex
\section{Experiments}

\input{Tables/PPL_Brier_Score}
\input{Tables/Paraphrase_Brier_Score}

\subsection{Datasets}
\label{sec:datasets}
\textbf{SNLI-VE} requires VLMs to identify whether the relationship between the given image premise and text hypothesis is entailment, neutral, or contradiction.
\textbf{Visual Commonsense Reasoning (VCR)} requires higher-order cognition and commonsense reasoning of VLMs. It provides an image and a question about certain objects in the image, along with four candidate answers, where the VLMs need to choose the correct answer.
We add rectangles of different colors to the image and indicate the corresponding object's index in the upper right corner of each rectangle to distinguish the objects.
\textbf{A-OKVQA} is an augmented successor of OK-VQA~\citep{marino2019ok} and requires a broad base of commonsense and world knowledge to answer questions. Four candidate answers are provided along with each question.
\textbf{Winoground (Wino.)} is proposed for measuring vision-linguistic compositional reasoning. It contains two images and two captions. The model needs to correctly match the captions to the images, but crucially, both captions contain an identical set of words, only in a different order.
\textbf{MMMU} is designed to evaluate VLMs on massive multi-discipline tasks demanding college-level subject knowledge and deliberate reasoning. Several candidate answers are provided along with each question.
\textbf{MathVista} focuses on mathematical reasoning in visual contexts.
We treat all datasets except for MMMU and MathVista as multiple-choice QA tasks. 
For evaluation:
\begin{itemize}
    \item For SNLI-VE, VCR, and A-OKVQA, we randomly select 1,000 samples from the validation set.
    \item For Winoground, we feed one image and two captions to the VLM, which must correctly identify the corresponding caption, using a total of 800 samples.
    \item For MMMU, we evaluate on the validation set, which contains 900 samples.
    \item For MathVista, we evaluate on the testmini set, which contains 1,000 samples.
\end{itemize}

\subsection{Implement Details}
\label{sec:implement}
We use InternVL-1.5~\cite{internvl} as the decomposer for decomposition and question paraphrasing. For decomposition, we employ few-shot prompting by randomly selecting four samples from SNLI-VE and ScienceQA, with manually written decomposition processes as guidance. The few-shot prompt for decomposition is provided in Table~\ref{tab:prompts_decomp}.
Only text is used in the few-shot prompt, without images. The decomposer determines the number of sub-questions needed. The few-shot prompt for the second-iteration decomposition is shown in Table~\ref{tab:prompts_decomp_second}
For paraphrasing, we use the same samples with manually written paraphrased questions. The few-shot prompt for paraphrasing is provided in Table~\ref{tab:prompts_paphrase}.
The remaining datasets are approached with a zero-shot strategy. We use OpenHermes-2.5-Mistral-7B\footnote{\url{https://huggingface.co/teknium/OpenHermes-2.5-Mistral-7B}} as the LLM Agent for reasoning. We evaluate three VLMs: LLaVA1.5-7B~\cite{liu2023llava}, Idefics2-8B~\cite{idefics2}, and InternVL~\cite{internvl}, all operating under a zero-shot setting across all datasets. Since all datasets are multiple-choice QA tasks or short answers, we use string matching for answer consistency.
For baseline threshold settings:
\begin{itemize}
    \item \textit{Perplexity of Direct Answer}: 1.10 for LLaVA1.5-7B, 1.25 for Idefics2-8B, and 1.40 for InternVL based on Table~\ref{Tab:ide_brier_ppl}.
    \item \textit{Generated Numerical Confidence}: We set the threshold to 80\%. If the generated confidence score exceeds 80\%, the reliability score is 1; otherwise, it is 0.
    \item \textit{Self-Consistency based on Paraphrase}: The number of inconsistent paraphrased-direct answer pairs is set to 0 for LLaVA1.5-7B and Idefics2-8B, and 2 for InternVL based on Table~\ref{tab:ide_brier_paraphrase}.
\end{itemize}

\begin{figure}[tb]
  \centering
  \includegraphics[width=0.95\linewidth]{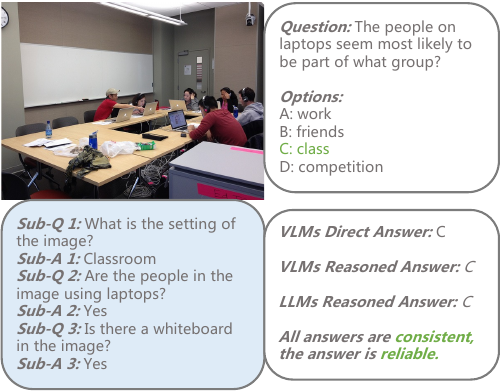}
  \caption{Example for the consistent situation. All answers are consistent, thus we assign the direct answer as reliable.}
  \label{fig:case_consist}
\end{figure}

\subsection{Evaluation Metric Selection}
\label{sec:metric_selection}
In our settings, we obtain binary reliability scores for each answer. We use the Brier Score~\cite{brier1950verification} and Effective Reliability~\cite{whitehead2022reliable} to evaluate the reliability measurement.
We do not use Expected Calibration Error (ECE)~\cite{guo2017calibration} because ECE is suitable for evaluating scores over a range of values. ECE relies on having a range of predicted probabilities to compare against actual accuracy. With only two reliability levels (0 or 1), there are no intermediate probabilities to assess the correlation.
We also find Coverage at Risk (C@R)~\cite{whitehead2022reliable} not applicable to our settings. C@R measures the \textbf{C}overage proportion of correctly answered questions if we tolerate an \textbf{R}\% of wrong answers by sorting predictions in descending order of score list and calculating coverage until the risk threshold is reached. C@R is not suitable for binary reliability scores because it relies on a range of reliability levels to sort and progressively evaluate predictions. With only binary scores, there is no meaningful way to sort the predictions by reliability. Consequently, C@R cannot provide a useful measure of performance in our setting.

\begin{figure}[h]
  \centering
  \includegraphics[width=0.85\linewidth]{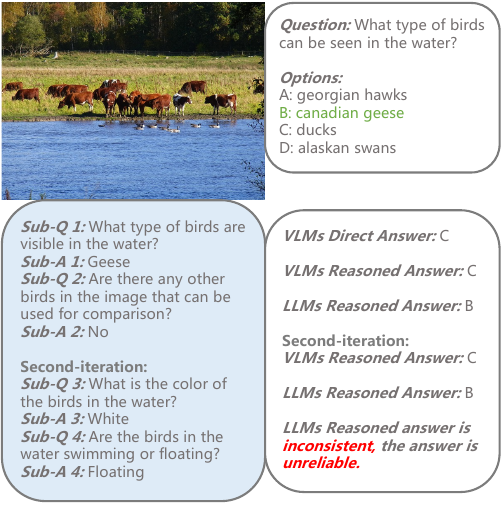}
\caption{Example for the inconsistent situation. The VLM's reasoned answer is consistent with the direct answer, while the LLM's reasoned answer is inconsistent. 
Both agents do not change their consistency check results.
We trust the LLM's consistency check results and assign the direct answer as unreliable.}
  \label{fig:case_inconsist}
\end{figure}

\begin{figure}[h]
  \centering
  \includegraphics[width=0.85\linewidth]{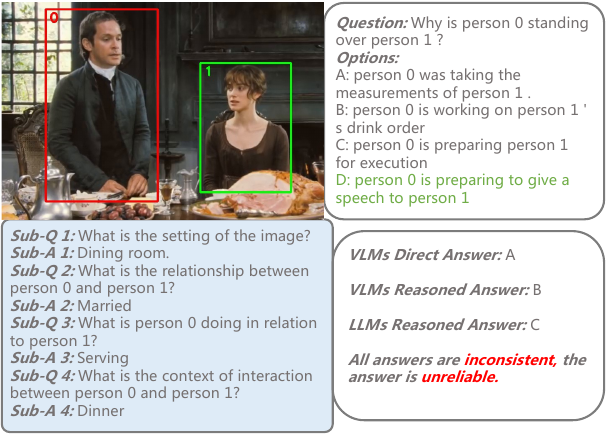}
\caption{Example for the inconsistent situation. All answers are inconsistent, while none of these answers are correct, indicating the VLMs do not understand the question well. We assign the direct answer as unreliable.}
  \label{fig:case_inconsist_wrong}
\end{figure}

\subsection{Case Study}
\label{sec:case_study}
Fig~\ref{fig:case_consist} shows an example from A-OKVQA where all answers are consistent, and we assign the direct answer as reliable.
Fig~\ref{fig:case_inconsist} shows an example from A-OKVQA where there is a contradiction between the consistency check results of the agents' reasoned answers and the direct answer. In this case, for the first sub-QA pair, the candidate VLM correctly identifies the birds as geese but fails to conduct correct reasoning over the decomposition process, deriving the same answer as the direct answer. Meanwhile, the LLM effectively utilizes the information from the decomposition. Both agents do not change their consistency check results.
As illustrated in Section~\ref{sec:consistency_compare}, we trust the LLM's consistency check results and assign the direct answer as unreliable. 
Fig~\ref{fig:case_inconsist_wrong} shows an example from VCR where all answers are inconsistent and incorrect, indicating that the VLMs do not understand the question well. We assign the direct answer as unreliable.

\input{Tables/Computation_Cost_VCR}

\subsection{Computational Costs Analysis}
\label{sec:comp_analysis}
\texttt{DeCC} requires multiple steps -- \textit{question decomposition, answering the decomposed questions, comparing consistency between the answers}. This raises concerns about \texttt{DeCC} being computationally more expensive than other approaches such as perplexity, generated numerical confidence scores, or self-consistency.
To address this concern, we provide detailed computational costs for \texttt{DeCC} and compare it with that of baselines.
Since the computational cost in \texttt{DeCC} is influenced by the number of sub-questions generated, we conduct this analysis on the VCR and AOKVQA, as these two datasets represent the median number of generated sub-questions among all benchmarks.
All statistics are computed using Idefics2-8B~\cite{idefics2} as the candidate VLM over a randomly chosen sample of 1,000 instances with a mean of three runs. For LLM Agent reasoning, we use vLLM~\cite{kwon2023efficient} for speeding up inference speed. Others are implemented using the Transformers toolkit\footnote{\url{https://huggingface.co/docs/transformers/en/index}} from Hugging Face without any speeding-up strategies. 
The consistency comparison stage requires only string comparisons, which is very fast (1 to 2 seconds for 1000 samples), so we do not list the time.
We report the computational costs analysis in Table~\ref{tab:computation_cost}.
For baselines such as perplexity of direct answer and generated confidence, the time cost is equal to direct answering, i.e., 0.22 seconds per sample for VCR and 0.16 seconds per sample for AOKVQA. 
For the baseline Self-Consistency based on Paraphrase, the consumed time is similar to the first iteration of DeCC, 4.8 seconds per sample for VCR and 3.9 seconds per sample for AOKVQA.
For the first iteration of DeCC, the most time-consuming stage is the Question Pre-Decomposition stage for generating the sub-questions, taking 3.96 seconds per sample for VCR and 3.36 seconds per sample for AOKVQA. However, this process needs to be performed only once per sample before evaluation and can then be reused to evaluate multiple VLMs, thus mitigating the impact on evaluation speed. The Sub-Question Answering and VLM Agent Reasoning + LLM Agent Reasoning takes 1.02 seconds per sample for VCR and 0.64 seconds per sample for AOKVQA. The second iteration of \texttt{DeCC} takes 5.22 seconds per sample for VCR and 4.57 seconds per sample for AOKVQA. However, the second decomposition is required only for samples where there is a disagreement between VLM Agent Consistency and LLM Agent Consistency from the first iteration, rather than for all samples. For example, only 25.3\% of samples in AOKVQA and 36.6\% of samples in VCR require the second iteration. Thus, the total time for the second iteration is significantly less than the first iteration. Overall the expected total time of \texttt{DeCC} Multi-Agent Setting taking into account both first and second round of decompositions is 6.89 seconds per sample for VCR and 5.16 seconds per sample for AOKVQA. Note that for VLM-Agent Consistency (2 Iterations) and LLM-Agent Consistency (2 Iterations), the expected total time is larger than the multi-agent setting because we need to do the second round of decomposition for all samples. With engineering optimizations and parallel processing, the computing time of \texttt{DeCC} can be further reduced.
Note that \texttt{DeCC} does not require any training or tuning of thresholds while other baselines such as perplexity, generated numerical confidence scores, or self-consistency require tuning certain thresholds, necessitating annotated datasets.
For example, we used 1,000 samples for threshold tuning in our experiments.
Data annotation is usually time-consuming and costly, averaging 10 seconds per sample\footnote{\href{https://www.cloudresearch.com/resources/blog/a-simple-formula-for-predicting-the-time-to-complete-a-study-on-mechanical-turk/}{CloudResearch: A Simple Formula for Predicting the Time to Complete a Study on Mechanical Turk}} with further filtering requests, and for specialized domains, this time can increase significantly, thus increasing the hidden costs. Compared with the data annotation, the time cost of \texttt{DeCC} is significantly lower.
In conclusion, despite the additional computation introduced by \texttt{DeCC} compared to baselines, our comprehensive experimental evaluation validates the effectiveness and practicality of \texttt{DeCC} across various benchmarks and VLMs.

\input{Tables/Consistency_algorithm}

\input{Tables/DecompositionPrompt}
\input{Tables/SecondLevelDecompositionPrompt}
\input{Tables/ParaphraseTemplate}

%% file: Tables/PPL_Brier_Score.tex
\begin{table*}[ht]
\centering
\resizebox{0.75\textwidth}{!}{\begin{tabular}{lccccccc}
\toprule
\textbf{Metric} & \textbf{SNLI}  & \textbf{VCR} & \textbf{A - OKVQA} & \textbf{Wino.} & \textbf{MMMU} & \textbf{MathVista} & \textbf{Mean} \\
\midrule
\textbf{LLaVA} \\
\text{Perplexity Threshold - 1.0} & 56.4 & 58.6 & 77.8 & 63.5 & 34.2 & 24.5 & 52.5 \\
\text{Perplexity Threshold - 1.05} & 56.4 & 47.4 & 36.0 & 58.1 & 31.5 & 24.6 & 42.3 \\
\text{Perplexity Threshold - 1.10} & 56.4 & 43.3 & 28.7 & 48.4 & 32.1 & 24.7 & 38.9 \\
\text{Perplexity Threshold - 1.15} & 56.2 & 41.9 & 25.1 & 41.3 & 35.2 & 25.1 & 37.5 \\
\text{Perplexity Threshold - 1.20} & 56.3 & 39.7 & 23.4 & 41.0 & 39.5 & 25.3 & 37.5 \\
\text{Perplexity Threshold - 1.25} & 55.7 & 38.2 & 22.8 & 39.3 & 42.4 & 25.9 & \textbf{37.4} \\
\hdashline
\textbf{Idefics2} \\
\text{Perplexity Threshold - 1.0} & 39.7 & 62.3 & 83.1 & 73.3 & 40.0 & 45.1 & 57.2 \\
\text{Perplexity Threshold - 1.05} & 59.1 & 33.3 & 22.6 & 32.6 & 36.6 & 31.6 & 35.9 \\
\text{Perplexity Threshold - 1.10} & 59.7 & 34.1 & 19.9 & 29.8 & 40.6 & 30.0 & \textbf{35.6} \\
\text{Perplexity Threshold - 1.15} & 60.1 & 36.5 & 18.5 & 27.9 & 43.9 & 31.0 & 36.3 \\
\text{Perplexity Threshold - 1.20} & 60.3 & 37.5 & 17.0 & 27.0 & 49.0 & 32.4 & 37.2 \\
\text{Perplexity Threshold - 1.25} & 60.2 & 38.0 & 16.6 & 26.6 & 53.0 & 35.0 & 38.2 \\

\hdashline

\textbf{InternVL} \\
\text{Perplexity Threshold - 1.0} & 70.2 & 71.1 & 88.5 & 80.2 & 43.6 & 55.2 & 68.1 \\
\text{Perplexity Threshold - 1.05} & 44.9 & 44.6 & 23.1 & 44.6 & 41.4 & 44.5 & 40.5 \\
\text{Perplexity Threshold - 1.10} & 38.8 & 38.0 & 17.9 & 37.1 & 39.2 & 40.8 & 35.3 \\
\text{Perplexity Threshold - 1.15} & 34.3 & 34.9 & 15.6 & 34.3 & 38.6 & 38.7 & 32.7 \\
\text{Perplexity Threshold - 1.20} & 31.8 & 32.5 & 14.1 & 31.3 & 38.9 & 35.4 & 30.7 \\
\text{Perplexity Threshold - 1.25} & 29.6 & 30.2 & 13.5 & 29.4 & 37.7 & 36.3 & 29.4 \\
\text{Perplexity Threshold - 1.30} & 28.3 & 29.1 & 12.7 & 27.5 & 36.6 & 36.1 & 28.4 \\
\text{Perplexity Threshold - 1.35} & 27.8 & 28.3 & 12.9 & 26.8 & 36.4 & 36.2 & 28.1 \\
\text{Perplexity Threshold - 1.40} & 28.0 & 27.5 & 12.1 & 24.0 & 37.3 & 36.5 & \textbf{27.6} \\
\bottomrule
\end{tabular}}
\caption{Brier Score using different threshold of perplexity on different VLMs.
Best results are in \textbf{bold}. All scores are in percentage. }
\label{Tab:ide_brier_ppl}
\end{table*}

%% file: Tables/Paraphrase_Brier_Score.tex
\begin{table*}[ht]
\centering
\resizebox{0.75\textwidth}{!}{\begin{tabular}{lccccccc}
\toprule
\textbf{Metric} & \textbf{SNLI}  & \textbf{VCR} & \textbf{A- OKVQA}& \textbf{Wino.} & \textbf{MMMU} & \textbf{MathVista} & \textbf{Mean} \\
\midrule
\textbf{LLaVA} \\
\text{Paraphrased Inconsistent - 0} & 38.5 & 32.8 & 19.0 & 40.5 & 39.1 & 35.6 & \textbf{34.3} \\
\text{Paraphrased Inconsistent - 1} & 39.5 & 34.1 & 19.2 & 37.1 & 46.6 & 44.1 & 36.8 \\
\text{Paraphrased Inconsistent - 2} & 41.2 & 36.4 & 19.9 & 37.6 & 50.0 & 49.7 & 39.1 \\

\hdashline
\textbf{Idefics2} \\
\text{Paraphrased Inconsistent - 0} & 59.1 & 31.6 & 16.3 & 28.9 & 41.6 & 40.8 & \textbf{36.4} \\
\text{Paraphrased Inconsistent - 1} & 60.4 & 31.5 & 15.8 & 28.0 & 46.4 & 41.4 & 37.3 \\
\text{Paraphrased Inconsistent - 2} & 61.1 & 31.6 & 16.1 & 27.8 & 47.4 & 43.9 & 38.0 \\

\hdashline

\textbf{InternVL} \\
\text{Paraphrased Inconsistent - 0} & 31.4 & 29.1 & 12.7 & 23.8 & 44.8 & 55.5 & 32.9 \\
\text{Paraphrased Inconsistent - 1} & 30.3 & 28.4 & 10.8 & 21.4 & 47.9 & 54.0 & 32.1 \\
\text{Paraphrased Inconsistent - 2} & 30.1 & 28.1 & 11.0 & 21.1 & 48.8 & 52.9 & \textbf{32.0} \\

\bottomrule
\end{tabular}}
\caption{Brier Score using different numbers of inconsistent paraphrased-direct answer pairs out of a total of 4 pairs. Best results are in \textbf{bold}. All scores are in percentage. }
\label{tab:ide_brier_paraphrase}
\end{table*}

%% file: Tables/Computation_Cost_VCR.tex
\begin{table*}[th]
\centering
\resizebox{\textwidth}{!}{
    \begin{tabular}{lccc}
        \toprule
        \multirow{2}{*}{\textbf{Method}} & \textbf{Sub-Questions/} & \textbf{Total Samples} & \textbf{Time per Sample} \\
        & \textbf{Paraphrased-Questions per Sample} & \textbf{Used} & \textbf{(Seconds)} \\
        \hline
        \textbf{VCR} & & & \\
        \hdashline
        \textbf{Baselines} & & & \\
        Perplexity of Direct Answer, Generated Confidence & N/a & 1000 & 0.22 \\
        Self-Consistency based on Paraphrase & 4 & 1000 & 4.80 \\
        \hdashline
        \textbf{\texttt{DeCC} Multi-Agent Consistency Stages} & & & \\
        Question Pre-Decomposition & 3.9 & 1000 & 3.96 \\
        Sub-Question Answering and VLM Agent Reasoning & 3.9 & 1000 & 0.84 \\
        LLM Agent Reasoning & 3.9 & 1000 & 0.18 \\
        Second Question Decomposition & 4.1 & 366 & 4.09 \\
        Second Sub-Question Answering and VLM Agent Reasoning & 4.1 & 366 & 0.93 \\
        Second LLM Agent Reasoning & 4.1 & 366 & 0.20 \\
        \hdashline
       \multirow{2}{*}{\textbf{\texttt{DeCC} Expected Value (Multi-Agent Consistency)}} & \multirow{2}{*}{$[1000 \times 3.9 + 366 \times 4.1] / 1000 = 5.4$} & \multirow{2}{*}{1000} & $[1000 \times (3.96 + 0.84 + 0.18) + $ \\
        & & & $366 \times (4.09 + 0.93 + 0.20)] / 1000 = 6.89$ \\
       \textbf{\texttt{DeCC} Expected Value (VLM-Agent Consistency (2 Iterations))} & 8.0 & 1000 & 9.82 \\
        \textbf{\texttt{DeCC} Expected Value (LLM-Agent Consistency (2 Iterations))} & 8.0 & 1000 & 10.20 \\
        \hline
        \hline
        \textbf{AOKVQA} & & & \\
        \hdashline
                \textbf{Baselines} & & & \\
        Perplexity of Direct Answer, Generated Confidence & N/a & 1000 & 0.16 \\
        Self-Consistency based on Paraphrase & 4 & 1000 & 3.90 \\
        \hdashline
        \textbf{\texttt{DeCC} Multi-Agent Consistency Stages} & & & \\
        Question Pre-Decomposition & 3.2 & 1000 & 3.36 \\
        Sub-Question Answering and VLM Agent Reasoning & 3.2 & 1000 & 0.54 \\
        LLM Agent Reasoning & 3.2 & 1000 & 0.10 \\
        Second Question Decomposition & 3.9 & 253 & 3.79 \\
        Second Sub-Question Answering and VLM Agent Reasoning & 3.9 & 253 & 0.66 \\
        Second LLM Agent Reasoning & 3.9 & 253 & 0.12 \\
        \hdashline
        \multirow{2}{*}{\textbf{\texttt{DeCC} Expected Value (Multi-Agent Consistency)}} & \multirow{2}{*}{$[1000 \times 3.2 + 253 \times 3.9] / 1000 = 4.19$} & \multirow{2}{*}{1000} & $[1000 \times (3.36 + 0.54 + 0.10) + $ \\
        & & & $253 \times (3.79 + 0.66 + 0.12)] / 1000 = 5.16$ \\
        \textbf{\texttt{DeCC} Expected Value (VLM-Agent Consistency (2 Iterations))} & 7.1 & 1000 & 8.35 \\
        \textbf{\texttt{DeCC} Expected Value (LLM-Agent Consistency (2 Iterations))} & 7.1 & 1000 & 8.57 \\
        \bottomrule
    \end{tabular}
}
\caption{Computational costs analysis on VCR and AOKVQA.}
\label{tab:computation_cost}
\end{table*}

%% file: Tables/Consistency_algorithm.tex
\begin{algorithm*}
\caption{Multi-Agent Consistency Comparison over Task Decomposition for Reliability Measurement}\label{alg:multi_agent_consistency}
\begin{algorithmic}[1]
\Require Question $Q$, Image $I$, Answer $A$, Decomposer, VLM for Evaluation, LLM for Reasoning
\Ensure Binary Reliability Score $\mathcal{R}$
\State Decomposer decomposes $Q$ into sub-questions
\State Generate sub-QA pairs by having VLM answer the sub-questions
\State Obtain $A_{V}^{'}$ and $A_{L}^{'}$ by reasoning over sub-QA pairs using VLM and LLM, respectively
\If{$A_{V}^{'}$ is consistent with $A$}
    \State $Cons_{V^{'}} \gets 1$
\Else
    \State $Cons_{V}^{'} \gets 0$
\EndIf
\If{$A_{L}^{'}$ is consistent with $A$}
    \State $Cons_{L}^{'} \gets 1$
\Else
    \State $Cons_{L}^{'} \gets 0$
\EndIf

\If{$Cons_{V}^{'} = Cons_{L}^{'}$}
    \State $\mathcal{R} \gets Cons^{'}$ \Comment{Direct determination}
\Else
    \State Perform second-iteration decomposition and generate new sub-QA pairs
    \State Obtain $A_{V}^{''}$ and $A_{L}^{''}$ by reasoning over all sub-QA pairs using VLM and LLM, respectively
    \If{$A_{V}^{''}$ is consistent with $A$}
        \State $Cons_{V}^{''} \gets 1$
    \Else
        \State $Cons_{V}^{''} \gets 0$
    \EndIf
    \If{$A_{L}^{''}$ is consistent with $A$}
        \State $Cons_{L}^{''} \gets 1$
    \Else
        \State $Cons_{L}^{''} \gets 0$
    \EndIf
    \If{$Cons_{V}^{''} = Cons_{L}^{''}$}
        \State $\mathcal{R} \gets Cons^{''}$ \Comment{Direct determination after second iteration}
    \Else
        \If{$Cons_{V}^{'} = Cons_{V}^{''}$ and $Cons_{L}^{'} = Cons_{L}^{''}$}
            \State $\mathcal{R} \gets Cons_{L}^{'}$ \Comment{LLM's consistency is used}
        \ElsIf{$Cons_{V}^{'} \neq Cons_{V}^{''}$ and $Cons_{L}^{'} \neq Cons_{L}^{''}$}
            \State $\mathcal{R} \gets Cons_{V}^{''}$ \Comment{VLM's consistency is used}
        \EndIf
    \EndIf
\EndIf

\end{algorithmic}
\end{algorithm*}

%% file: Tables/DecompositionPrompt.tex
\begin{table*}
    \centering
    \small
    \begin{tabular}{lp{13cm}}
        \toprule
        & \textbf{Few-Shot Prompt for Decomposition} \\
        \midrule
        & {Given an image and an associated main question, design pre-questions that focus on important contextual information in the image useful for answering the main question. Pre-questions should provide clues to answer the main question. Each pre-question should be short and easy to understand. Pre-questions should focus on context visual clues of the image. Pre-questions should provide clues to answer the main question.} \\
        \\
        & \textbf{Example scenario to illustrate the expected interaction pattern:} \\
        & Main Question: Is this statement entailment, neutral or contradiction based on the image? Statement: `A professor is late to class' Options: A: entailment, B: neutral, C: contradiction. \\
        & Pre-question 1: Is there a person in the image wearing clothing typically associated with a professor? \\
        & Pre-question 2: Is the person in the image displaying any behavior that could be interpreted as being late to class, such as being out of breath or looking at a clock? \\
        & Pre-question 3: Is there a classroom setting in the image, such as desks or a blackboard? \\
        \\
        & \textbf{Example scenario to illustrate the expected interaction pattern:} \\
        & Context: Below is a food web from a tundra ecosystem in Nunavut, a territory in Northern Canada. A food web models how the matter eaten by organisms moves through an ecosystem. The arrows in a food web represent how matter moves between organisms in an ecosystem. Main Question: Based on the arrows, which of the following organisms is a decomposer? Choices: A: mushroom, B: lichen \\
        & Pre-question 1: Does the mushroom eat any other organisms in the food web? \\
        & Pre-question 2: Does the lichen eat any other organisms in the food web? \\
        & Pre-question 3: Does the lichen produce any material that other organisms can use? \\
        & Pre-question 4: Does the mushroom produce any material that other organisms can use? \\
        & Pre-question 5: Does a decomposer produce any material that other organisms can use? \\
        \\
        & \textbf{Example scenario to illustrate the expected interaction pattern:} \\
        & Main Question: Is this statement entailment, neutral or contradiction based on the image? Statement: `Two children play in the park.' Options: A: entailment, B: neutral, C: contradiction. \\
        & Pre-question 1: Are there any children in the image? \\
        & Pre-question 2: Are the two children playing in the park? \\
        \\
        & \textbf{Example scenario to illustrate the expected interaction pattern:} \\
        & User: Context: Use the graph to answer the question below. Main Question: Which month has the highest average precipitation in Santiago? Choices: A: March, B: October, C: June \\
        & Pre-question 1: What kind of graph is shown? \\
        & Pre-question 2: Does the graph show the average precipitation for each month in Santiago? \\
        & Pre-question 3: For which month is the bar highest in the graph? \\
        \bottomrule
    \end{tabular}
    \caption{Few-Shot Prompt for Decomposition.}
    \label{tab:prompts_decomp}
\end{table*}

%% file: Tables/SecondLevelDecompositionPrompt.tex
\begin{table*}
    \centering
    \small
    \begin{tabular}{lp{13cm}}
        \toprule
        & \textbf{Few-Shot Prompt for Second-Iteration Decomposition} \\
        \midrule
        & You will be given an image and an associated main question, and some sub-question-answer pairs. However, these sub-questions might not be sufficient to answer the main question due to lack of detail or conflicting answers. You need to design additional sub-questions that focus on important contextual information in the image useful for answering the main question. Each pre-question should be short, easy to understand, and provide clues to answer the main question. \\
        \\
        & \textbf{Example scenario to illustrate the expected interaction pattern:} \\
        & Main Question: Is this statement entailment, neutral, or contradiction based on the image? Statement: `A professor is late to class' Options: A: entailment, B: neutral, C: contradiction. \\
        & Sub-questions and answers: \\
        & Sub-question 1: Is there a person in the image wearing clothing typically associated with a professor? \\
        & Sub-answer 1: Yes. \\
        & Sub-question 2: Is the person in the image displaying any behavior that could be interpreted as being late to class, such as being out of breath or looking at a clock? \\
        & Sub-answer 2: No. \\
        & Sub-question 3: Is there a classroom setting in the image, such as desks or a blackboard? \\
        & Sub-answer 3: Yes. \\
        & Your return: \\
        & Additional Sub-question 1: What is the person's age in the image? \\
        & Additional Sub-question 2: Is the person more likely to be a student or a professor? \\
        & Additional Sub-question 3: Is the person holding any books or papers? \\
        \\
        & \textbf{Example scenario to illustrate the expected interaction pattern:} \\
        & Context: Below is a food web from a tundra ecosystem in Nunavut, a territory in Northern Canada. A food web models how the matter eaten by organisms moves through an ecosystem. The arrows in a food web represent how matter moves between organisms in an ecosystem. Main Question: Based on the arrows, which of the following organisms is a decomposer? Choices: A: mushroom, B: lichen. \\
        & Sub-questions and answers: \\
        & Sub-question 1: Does the mushroom eat any other organisms in the food web? \\
        & Sub-answer 1: Yes. \\
        & Sub-question 2: Does the lichen eat any other organisms in the food web? \\
        & Sub-answer 2: No. \\
        & Sub-question 3: Does the lichen produce any material that other organisms can use? \\
        & Sub-answer 3: Yes. \\
        & Sub-question 4: Does the mushroom produce any material that other organisms can use? \\
        & Sub-answer 4: No. \\
        & Sub-question 5: Does a decomposer produce any material that other organisms can use? \\
        & Sub-answer 5: Yes. \\
        & Your return: \\
        & Additional Sub-question 1: Is there any arrow pointing towards the mushroom? \\
        & Additional Sub-question 2: Is there any arrow pointing towards the lichen? \\
        & Additional Sub-question 3: What is the mushroom's role in the food web? \\
        & Additional Sub-question 4: What is the lichen's role in the food web? \\
        \bottomrule
    \end{tabular}
    \caption{Few-Shot Prompt for Second-Iteration Decomposition.}
    \label{tab:prompts_decomp_second}
\end{table*}

%% file: Tables/ParaphraseTemplate.tex
\begin{table*}
    \centering
    \small
    \begin{tabular}{lp{13cm}}
        \toprule
        & \textbf{Few-Shot Prompt for Paraphrase} \\
        \midrule
        & Your goal is to paraphrase the given question into 4 questions. Each question should only change the wording of the original question slightly or just replace a few words. The questions should be easy to understand and should not change the meaning of the original question. If the questions come with some choices, you should not change these choices. \\
        \\
        & \textbf{Example scenario to illustrate the expected interaction pattern:} \\
        & Main Question: Is this statement entailment, neutral, or contradiction based on the image? Statement: `A professor is late to class' Options: A: entailment, B: neutral, C: contradiction. \\
        & Paraphrased question 1: Is this statement entailment, neutral, or contradiction based on the image? Statement: `A teacher is late to class' Options: A: entailment, B: neutral, C: contradiction. \\
        & Paraphrased question 2: Is this statement entailment, neutral, or contradiction based on the image? Statement: `A professor is tardy to class' Options: A: entailment, B: neutral, C: contradiction. \\
        & Paraphrased question 3: Is this statement entailment, neutral, or contradiction based on the image? Statement: `A professor is not on time for class' Options: A: entailment, B: neutral, C: contradiction. \\
        & Paraphrased question 4: Is this statement entailment, neutral, or contradiction based on the image? Statement: `A teacher is not punctual for class' Options: A: entailment, B: neutral, C: contradiction. \\
        \\
        & \textbf{Example scenario to illustrate the expected interaction pattern:} \\
        & Context: Below is a food web from a tundra ecosystem in Nunavut, a territory in Northern Canada. A food web models how the matter eaten by organisms moves through an ecosystem. The arrows in a food web represent how matter moves between organisms in an ecosystem. Main Question: Based on the arrows, which of the following organisms is a decomposer? Choices: A: mushroom, B: lichen \\
        & Paraphrased question 1: Based on the arrows, which of these choices is a decomposer? Choices: A: mushroom, B: lichen \\
        & Paraphrased question 2: Based on the arrows, which of the following is a decomposer? Choices: A: mushroom, B: lichen \\
        & Paraphrased question 3: Which of the following is a decomposer based on the arrows? Choices: A: mushroom, B: lichen \\
        & Paraphrased question 4: Which is a decomposer based on the figure? Choices: A: mushroom, B: lichen \\
        \\
        & \textbf{Example scenario to illustrate the expected interaction pattern:} \\
        & Main Question: Is this statement entailment, neutral, or contradiction based on the image? Statement: `Two children play in the park.' Options: A: entailment, B: neutral, C: contradiction. \\
        & Paraphrased question 1: Is this statement entailment, neutral, or contradiction based on the image? Statement: `Two kids play in the park.' Options: A: entailment, B: neutral, C: contradiction. \\
        & Paraphrased question 2: Is this statement entailment, neutral, or contradiction based on the image? Statement: `Two children are playing in the park.' Options: A: entailment, B: neutral, C: contradiction. \\
        & Paraphrased question 3: Is this statement entailment, neutral, or contradiction based on the image? Statement: `Two kids are playing in the park.' Options: A: entailment, B: neutral, C: contradiction. \\
        & Paraphrased question 4: Is this statement entailment, neutral, or contradiction based on the image? Statement: `There are two children playing in the park.' Options: A: entailment, B: neutral, C: contradiction. \\
        \\
        & \textbf{Example scenario to illustrate the expected interaction pattern:} \\
        & User: Context: Use the graph to answer the question below. Main Question: Which month has the highest average precipitation in Santiago? Choices: A: March, B: October, C: June \\
        & Paraphrased question 1: Which month has the highest average rainfall in Santiago? Choices: A: March, B: October, C: June \\
        & Paraphrased question 2: Which month's precipitation is the highest in Santiago? Choices: A: March, B: October, C: June \\
        & Paraphrased question 3: Which month has the most precipitation in Santiago? Choices: A: March, B: October, C: June \\
        & Paraphrased question 4: Which month has the most rainfall in Santiago? Choices: A: March, B: October, C: June \\
        \\
        & Note: Return the paraphrased questions. For each paraphrased question, you should return the entire set of choices as well. \\
        \bottomrule
    \end{tabular}
    \caption{Few-Shot Prompt for Paraphrase.}
    \label{tab:prompts_paphrase}
\end{table*}